\pdfoutput=1

\documentclass[11pt]{article}

\usepackage{EMNLP2022}

\usepackage{times}
\usepackage{latexsym}

\usepackage[T1]{fontenc}

\usepackage[utf8]{inputenc}

\usepackage{microtype}

\usepackage{inconsolata}

\usepackage{balance}
\usepackage{url}
\usepackage{graphicx}
\usepackage{amsmath,amssymb,amsfonts}
\usepackage{algorithmic}
\usepackage{textcomp}
\usepackage{xcolor}
\usepackage{tikz}
\usetikzlibrary{topaths,calc}
\usepackage[ruled]{algorithm2e}
\usepackage{subfigure}
\usepackage{booktabs}
\usepackage{balance}
\usepackage{colortbl}
\usepackage{enumitem}
\usepackage{multirow}
\usepackage[normalem]{ulem}

\usepackage{etoolbox}

\newcommand{\revisejie}[1]{#1}

\title{Understanding Jargon: Combining Extraction and Generation for Definition Modeling}

\author{Jie Huang$^{1}$ $\quad$ Hanyin Shao$^{1}$ $\quad$ Kevin Chen-Chuan Chang$^{1}$ \\ \textbf{Jinjun Xiong$^{2}$ $\quad$ Wen-mei Hwu$^{1,3}$} \\
 $^1$University of Illinois at Urbana-Champaign, USA \\
 $^2$University at Buffalo, USA \\
 $^3$NVIDIA, USA \\
 \texttt{\{jeffhj, hanyins2, kcchang, w-hwu\}@illinois.edu} \\
 \texttt{jinjun@buffalo.edu}
}

\begin{document}
\maketitle
\begin{abstract}
Can machines know what \textit{twin prime} is? From the composition of this phrase, machines may guess \textit{twin prime} is a certain kind of prime, but it is still difficult to deduce exactly what \textit{twin} stands for without additional knowledge. Here, \textit{twin prime} is a jargon-- a specialized term used by experts in a particular field. Explaining jargon is challenging since it usually requires domain knowledge to understand. Recently, there is an increasing interest in extracting and generating definitions of words automatically. However, existing approaches, either extraction or generation, perform poorly on jargon. In this paper, we propose to combine extraction and generation for jargon definition modeling: first extract self- and correlative definitional information of target jargon from the Web and then generate the final definitions by incorporating the extracted definitional information. Our framework is remarkably simple but effective: experiments demonstrate our method can generate high-quality definitions for jargon and outperform state-of-the-art models significantly, e.g., BLEU score from 8.76 to 22.66 and human-annotated score from 2.34 to 4.04.\footnote{Code and data are available at \url{https://github.com/jeffhj/CDM}.}
\end{abstract}

\section{Introduction}

Jargons are specialized terms associated with a particular discipline or field. To understand jargons, a straightforward approach is to read their definitions, which are highly summarized sentences that capture the main characteristics of them. For instance, given jargon \textit{twin prime}, people can know its meaning by reading its definition: ``\textit{A twin prime is a prime number that is either 2 less or 2 more than another prime number.}''

Recently, acquiring definitions of words/phrases automatically has aroused increasing interest. 
There are two main approaches: 
\emph{extractive}, corresponding to \emph{definition extraction}, where definitions are extracted from existing corpora automatically \cite{anke2018syntactically,veyseh2020joint,kang2020document}; and \emph{abstractive}, corresponding to \emph{definition generation}, where definitions are generated conditioned with the target words/phrases and the contexts in which they are used \cite{noraset2017definition,gadetsky2018conditional,bevilacqua2020generationary,august-2022-definition-complexity,gardner2022definition}. 

In this paper, we study \textbf{jargon definition modeling}, which aims to acquire definitions for jargon automatically. 
Jargon definition modeling is important since definitions of jargon are less likely to be organized in an existing dictionary/encyclopedia and such terms are difficult for non-experts to understand without explanations \cite{bullock2019jargon}. {This is particularly true for new jargon from fast-advancing fields.}
For instance, neither Oxford dictionary \cite{butterfield2016dictionary} nor Wikipedia\footnote{\url{https://en.wikipedia.org}} includes \textit{few-shot learning}-- an important setup in machine learning.

\begin{figure*}[tp!]
\centerline{\includegraphics[width=0.9\linewidth]{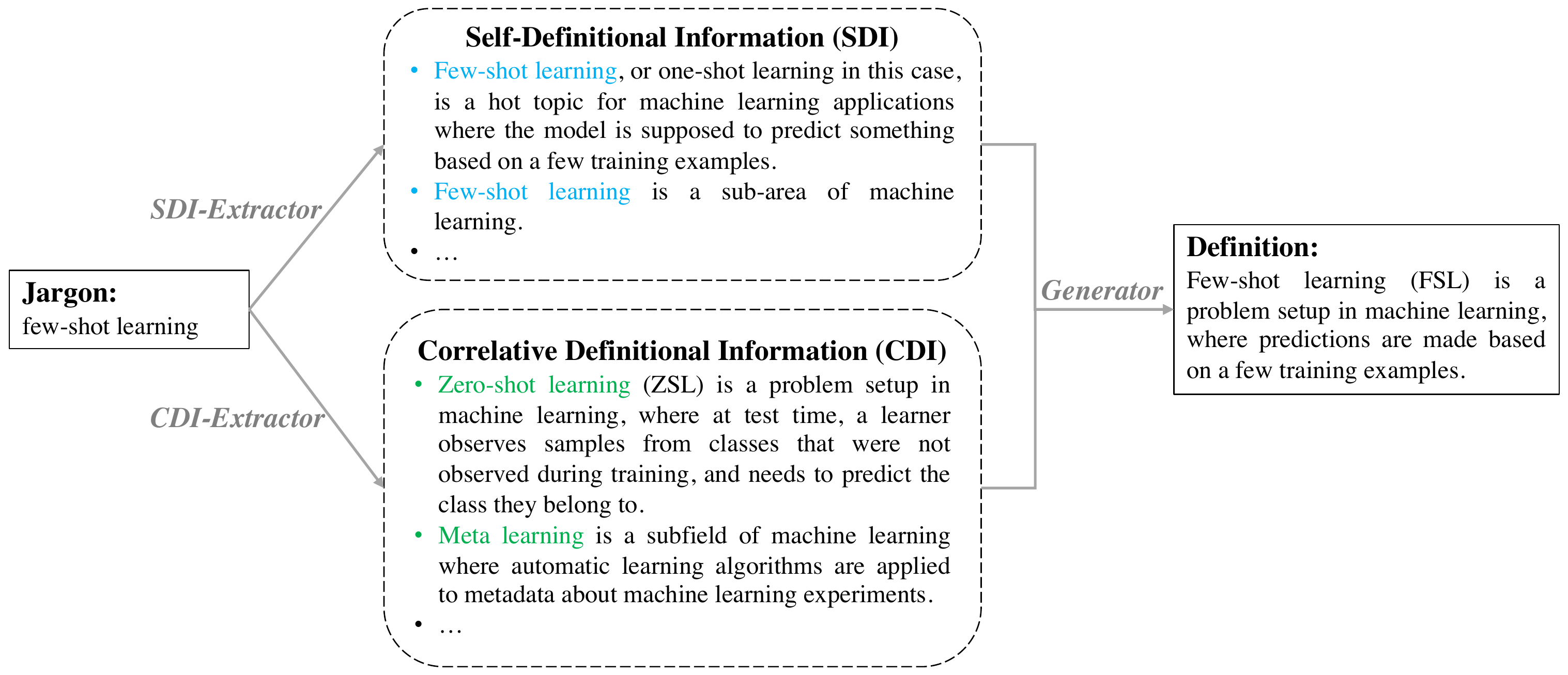}}
\vspace{-1mm}
\caption{The overview of the proposed framework. In this example, the definition of \textit{few-shot learning} is generated based on both the SDI (e.g., ``\textit{predictions are made based on a few training examples}'') and the CDI (e.g., ``\textit{is a problem setup in machine learning}'').}
\vspace{-2mm}
\label{fig:CDM}
\end{figure*}

However, to acquire definitions for jargon, both extractive and abstractive approaches may fail. 
Extracting high-quality definitions would be difficult due to the incompleteness and low quality of data sources (this issue is more serious for jargon since jargon is usually less frequently used %
than general words/phrases). E.g., a good definition may not be available in the corpus; even if it existed, it might be difficult to select from a large set of candidate sentences \cite{kang2020document}. 
Generating definitions for jargon would be challenging since jargons are usually technical terms that need domain knowledge to understand, while the contexts in which they are used cannot provide sufficient knowledge. For instance, it is almost impossible for a model to generate the definition for \textit{twin prime} only with context ``\textit{proof of this conjecture would also imply the existence of an infinite number of \uline{twin primes}}'' since the context does not explain \textit{twin prime}, and the specific meaning is difficult to infer from the surface form, leading to \textit{hallucinations}, i.e., generating irrelevant or contradicted facts \cite{bevilacqua2020generationary}.
Consequently, existing models designed for general words/phrases perform poorly on jargon. In our evaluation (Tables \ref{table:human_evaluation} and \ref{table:example}), we find most definitions produced by the state-of-the-art model contain wrong information.

Fortunately, definition extraction and definition generation can complement each other naturally. On one hand, definition generator has the potential to help the extractor by refining and synthesizing the extracted definitions; therefore, the extracted sentences are not required to be perfect definitions of the target jargon. 
On the other hand, definition extractor can retrieve useful definitional information as knowledge for the generator to produce definitions of jargon.
However, surprisingly, existing works are either extractive or abstractive, even do not connect and compare them.  

Therefore, in this work, we propose to combine definition extraction and definition generation for jargon definition modeling. We achieve this by introducing a framework consisting of two processes: \emph{extraction}, where definitional information of jargon is extracted from the Web; and \emph{generation}, where the final definition is generated with the help of the extracted definitional information. 

We build models for extraction and generation based on Pre-Trained Language Models \cite{devlin2019bert,lewis2020bart,NEURIPS2020_1457c0d6}.
Specifically, 
for extraction, we propose a BERT-based definition extractor to extract \emph{self-definitional information} (i.e., definitional sentences of the target jargon).
We also suggest that related terms can help define the target jargon and leverage Wikipedia as the external knowledge source to retrieve \emph{correlative definitional information} (i.e., definitions of related terms).
For generation, we design a BART-based definition generator to produce the final definition by incorporating the extracted knowledge. An example is shown in Figure \ref{fig:CDM}.

Our framework for jargon definition modeling is remarkably simple that can easily be further expanded by leveraging more advanced language models, e.g., we can replace the BART generator with larger models such as Meta OPT \cite{zhang2022opt} with a simple modification. 
Besides, since our framework does not require a domain-specific corpus or ontology like the ones used in \citet{vanetik2020automated,liu2021graphine}, it is easy to apply to a variety of domains.
Experimental results on four datasets demonstrate our \textbf{\emph{simple}} model outperforms state-of-the-art models \textbf{\emph{significantly}} (e.g., BLEU score \uline{from 8.76 to 22.66}, human-annotated score \uline{from 2.34 to 4.04}).

Our contributions are summarized as follows:
\begin{itemize}[noitemsep,nolistsep,leftmargin=*]
\item We report the first attempt to connect and combine definition extraction and definition generation.
\item We introduce jargon definition modeling and solve it by incorporating both self- and correlative definitional information of jargon.
\item Experimental results show that our simple model substantially outperforms SOTA models for definition modeling.
\item We publish several datasets, along with definitions (e.g., of \texttt{\textasciitilde 75,600} computer science terms) generated by our proposed model.
\end{itemize}

\section{Related Work}

{\flushleft \textbf{Definition Extraction}.}
Definition extraction, which aims to extract definitions from corpus automatically, has been studied for a long period.
Existing works for definition extraction can be roughly divided into three categories: 1) \emph{rule-based}, which extracts definitions with defined linguistic rules and templates \cite{klavans2001evaluation,cui2004unsupervised,fahmi2006learning}; 2) \emph{machine learning-based}, which extracts definitions by statistical machine learning with carefully designed features \cite{westerhout2009definition,jin2013mining}; 3) \emph{deep learning-based}, the state-of-the-art approach for definition extraction, which is based on deep learning models such as CNN, LSTM, and BERT \cite{anke2018syntactically,veyseh2020joint,kang2020document,vanetik2020automated}.

{\flushleft \textbf{Definition Generation}.}
Definition generation, or definition modeling, has aroused increasing interest in recent years.
The first study on definition generation was presented in \citet{noraset2017definition}, which aims to generate definitions of words with word embeddings.
Later works on definition generation put more emphasis on generating definitions of words/phrases with given contexts \cite{gadetsky2018conditional,ishiwatari2019learning,washio2019bridging,mickus2019mark,li2020explicit,reid2020vcdm,bevilacqua2020generationary,huang2021definition}. 
For example, \citet{bevilacqua2020generationary} apply pre-trained BART \cite{lewis2020bart} for definition generation with a simple context encoding scheme. \citet{huang2021definition} employ three T5 models \cite{raffel2020exploring} for definition generation with a re-ranking mechanism to model specificity of definitions. \citet{liu2021graphine} study the graph-aware definition modeling problem by incorporating biomedical ontology. \revisejie{\citet{august-2022-definition-complexity} study the problem of generating deﬁnitions of scientific and medical terms with varying complexity. \citet{huang2022open} propose to generate definitional-like sentences to describe relations between entities.}
There are also recent works on definition modeling for other languages, e.g., Chinese, by incorporating the special properties of the specific language \cite{yang2020incorporating,zheng2021decompose}.

However, although definition extraction and definition generation are quite relevant tasks, surprisingly, existing works do not connect and compare them. In this work, we report the first attempt to combine them.

\section{Methodology}

Our framework for jargon definition modeling consists of two processes: \emph{extraction}, which extracts self- and correlative definitional information of the target jargon from the Web;
and \emph{generation}, which generates the final definition by incorporating the extracted definitional information. The overview of the framework is shown in Figure \ref{fig:CDM}.

\subsection{Extraction}
\label{sec:extraction}

\subsubsection{Self-Definitional Information}
\label{sec:sdi}

Since jargons are specialized terms used in a particular field, to understand jargon, we need background knowledge of jargon.
To acquire useful information for defining jargon, it is natural to refer to definitional sentences containing the target jargon, named \textbf{\emph{Self-Definitional Information (SDI)}}.
We achieve SDI by first extracting sentences containing the target jargon from the Web (more details are in Section \ref{sec:data}) and then using a classifier to rank the extracted sentences.

To build the classifier, we apply the BERT model \cite{devlin2019bert}, which has achieved excellent results on various text classification tasks.
We adopt a simple encoding scheme, which is ``\textit{[CLS] $jargon$ [DEF] $sentence$}'', e.g., ``\textit{[CLS] machine learning [DEF] machine learning is the study of computer algorithms that improve automatically through experience and by the use of data.}'' 
The final hidden state of the first token \texttt{[CLS]} is used as the representation of the whole sequence and a classification layer is added.
After fine-tuning on the jargon-sentence pairs, the model has a certain ability to distinguish whether the sentence contains representative definitional information of the target jargon.
SDI is then obtained as the top definitional sentences by ranking the sentences according to the confidence of the prediction.
We refer to this model as \textbf{SDI-Extractor}.

\subsubsection{Correlative Definitional Information}

To explain a jargon, in addition to utilizing SDI, we can also refer to the definitions of its related terms, i.e., \textbf{\emph{Correlative Definitional Information (CDI)}}. 
For instance, to define \textit{few-shot learning}, we can incorporate definitions of \textit{zero-shot learning} and \textit{meta learning}, with which we can know the meaning of ``shot'' and ``learning'' and may define \textit{few-shot learning} similarly to \textit{zero-shot learning}. 

To get related terms and their definitions, we leverage Wikipedia as the external knowledge source, which covers a wide range of domains and contains high-quality definitions for a large number of terms. Specifically, we follow the \textit{core-fringe} notion in \citet{huang2021measuring}, where \textit{core terms} are terms that have corresponding Wikipedia pages, and \textit{fringe terms} are ones that are not associated with a Wikipedia page.
For each jargon, we treat it as query to retrieve the most relevant core terms via document ranking based on Elasticsearch \cite{gormley2015elasticsearch}, and extract first sentences on the corresponding Wikipedia pages as the definitions of related terms.
We refer to this model as \textbf{CDI-Extractor}.

\subsection{Generation}
\label{sec:generation}

After extraction, we acquire the self- and correlative definitional information of jargon. This kind of information captures important characteristics of jargon and can be further refined and synthesized into the final definition by a definition generator.

Definition generation can be formulated as a conditioned sentence generation task-- generating a coherent sentence to define the target jargon. Formally, we apply the standard sequence-to-sequence formulation: given jargon $x$, combining with the extracted sentences $\mathcal{S}_s$ (for SDI) and $\mathcal{S}_c$ (for CDI), the probability of the generated definition $d$ is computed auto-regressively:
\begin{equation*}
P(d|x,\mathcal{S}_{s}, \mathcal{S}_{c}) = \prod_{i=1}^{m} P(d_i|d_{0:i-1}, x, \mathcal{S}_{s}, \mathcal{S}_{c}),
\end{equation*}
where $m$ is the length of $d$, $d_i$ is the $i$th token of $d$, and $d_0$ is a special start token.

Following \citet{bevilacqua2020generationary}, to build the generator, we employ BART \cite{lewis2020bart}, a pre-trained transformer-based encoder-decoder model that can be fine-tuned to perform specific conditional language generation tasks with specific training input-output pairs.
Different from existing works \cite{gadetsky2018conditional,ishiwatari2019learning,bevilacqua2020generationary} which aim to learn to define a word/phrase in a given context, we propose to learn to define a jargon using the extracted knowledge. 
To be specific, we aim to fine-tune the BART model to generate the definition of the target jargon based on the surface name of the jargon and the extracted definitional information. 

To apply the BART model, for a target jargon, we adopt the following encoding scheme: ``\textit{$jargon$ [DEF] $sent_1$ [SEP] $sent_2$ ... [SEP] $sent_{k}$ [DEF] $sent'_1$ [SEP] $sent'_2$ ... [SEP] $sent'_{k'}$}'', where $sent_i$ and $sent'_i$ are the $i$th sentences ranked by \emph{SDI-Extractor} and \emph{CDI-Extractor}, respectively. We fine-tune BART to produce the ground-truth definition conditioned with the encoded input. 

After training, given a new jargon, we get corresponding SDI and CDI according to Section \ref{sec:extraction}. 
We encode the jargon and the top $k$ ranked sentences of SDI and top $k'$ ranked sentences of CDI as described above and use the generator to produce the final definition.
We refer to this model as \textbf{CDM-S$k$,C$k'$}, i.e., \textbf{C}ombined \textbf{D}efinition \textbf{M}odeling.

Here we would like to mention that our combined definition modeling framework is modular and can be applied to different extractor-generator combinations commonly proposed for definition extraction/generation, which means that the proposed framework can improve the performance for a variety of definition modeling systems. For instance, we can replace the BART generator with GPT-2/3 generator \cite{radford2019language,NEURIPS2020_1457c0d6} or DMAS \cite{huang2021definition} by simply modifying the encoding scheme. 

\section{Experiments}

\subsection{Datasets}

\label{sec:data}

Existing datasets for definition modeling are mainly for general words/phrases. 
In this paper, we build several datasets (\textbf{UJ-CS}, \textbf{UJ-Math}, \textbf{UJ-Phy}) for jargon based on Wikipedia and CFL \cite{huang2021measuring}.
Compared to general words/phrases, jargons are less ambiguous but more specialized, i.e., a jargon usually only has one meaning, but it requires domain knowledge to understand. \revisejie{We also conduct experiments on the dataset (\textbf{Sci\&Med}) provided in \citet{august-2022-definition-complexity}, which contains definitions of scientific and medical terms derived from Wikipedia science glossaries and MedQuAD \cite{ben2019question}.}

{\flushleft \textbf{Definition Extraction}.} 
We build a dataset for jargon definition extraction with Wikipedia.
We first collect jargons with Wikipedia Category. Specifically, we traverse from three root categories, including \textit{Category:Subfields of computer science}\footnote{\url{https://en.wikipedia.org/wiki/Category:Subfields_of_computer_science}}, \textit{Category:Fields of mathematics}\footnote{\url{https://en.wikipedia.org/wiki/Category:Fields_of_mathematics}}, and \textit{Category:Subfields of physics
}\footnote{\url{https://en.wikipedia.org/wiki/Category:Subfields_of_physics}}, and collect pages at the first three levels of the hierarchies. 
For each page, we process the title with lemmatization as the jargon, extract the first sentence in the summary section as the corresponding definition, and sample $\leq5$ sentences containing the target jargon from other sections as negatives (they are less likely to be definitional sentences).
We filter out jargons with surface name frequency $<5$ in the arXiv corpus\footnote{\url{https://www.kaggle.com/Cornell-University/arxiv}} 
(to filter out some noisy phrases, e.g., \textit{List of artificial intelligence projects}).
The dataset contains 26,559 positive and 121,975 negative examples, and the train/valid/test split is 0.8/0.1/0.1.

{\flushleft \textbf{Definition Generation}.} 
Following \cite{huang2021measuring}, we focus on generating definitions for jargon in three fields: computer science (\textbf{UJ-CS}), mathematics (\textbf{UJ-Math}), and physics (\textbf{UJ-Phy}). We collect jargons in two ways. For computer science,  
we collect jargons (author-assigned keywords) by web scraping from Springer publications on computer science. We filter out jargons with frequency $<5$.
For mathematics and physics, we collect jargons with the CFL model proposed in \citet{huang2021measuring}. Specifically, we collect terms with domain relevance score $>0.5$ as jargons.
For each jargon in the list, URLs of the top 20 results from Google search are visited. Then the sentences containing the target jargon are extracted. 
For training and evaluation, we only keep jargons that have a Wikipedia page and extract the first sentence on each page as the ground-truth definition. Table~\ref{table:dataset} summarizes the statistics of the data.

\begin{table}[tp!]
    \small
    \begin{center}
    \setlength\tabcolsep{5pt}
    \begin{tabular}{l|c|r|r|r}
        \toprule
        \textbf{Data} & \textbf{Source of Jargon} &  \textbf{Train} & \textbf{Valid} & \textbf{Test} \\
        \midrule
        \textbf{UJ-CS} & Springer & 11,738 & 1,671 & 3,349 \\
        \hline
        \textbf{UJ-Math} & CFL & 4,247 & 583 & 1,019 \\
        \hline
        \textbf{UJ-Phy} & CFL  & 4,157 & 573 & 1,026 \\
        \bottomrule
    \end{tabular}
    \end{center}
    \vspace{-1mm}
    \caption{The statistics of the data.}
    \vspace{-3mm}
    \label{table:dataset}
\end{table}

\subsection{Experimental Setup}

{\flushleft \textbf{Baselines}.}
For extraction, we compare SDI-Extractor with a CNN baseline and a CNN-BiLSTM baseline proposed in \citet{anke2018syntactically}. 
Here we should mention that the more recent models \cite{veyseh2020joint,kang2020document} cannot be compared directly since these works focus on a fine-grained sequence labeling task, where the training data also requires additional labeling. Besides, extraction is not the focus of this paper; therefore, we put more emphasis on the evaluation for generation.
For generation, we evaluate on the following models:
\begin{itemize}[noitemsep,nolistsep,leftmargin=*]
\item \textbf{Gen (w/o context)}: A simple version of Generationary \cite{bevilacqua2020generationary}, 
where BART \cite{lewis2020bart} is fine-tuned on jargon-definition pairs.
\item \textbf{Gen (w/ context)}: Generationary with a sentence containing the target jargon as context, 
where BART is fine-tuned on context-definition pairs. 
\item \textbf{DMAS} \cite{huang2021definition}: A definition modeling model with three T5 \cite{raffel2020exploring}, where a re-ranking mechanism is included to model the specificity of definitions. Context is given by a sentence containing the target jargon.
\item \revisejie{\textbf{BART NO SD and BART SD}: For the \textit{Sci\&Med} dataset \cite{august-2022-definition-complexity}, we also compare with the two best methods introduced in their paper: BART SD, where BART is fine-tuned with the term question, e.g., \textit{What is (are) carbon nanotubes?}, concatenated with the supporting document; and BART NO SD, where BART is fine-tuned with just the question and definition, without the support documents.}
\item \textbf{Extractive}: An extractive baseline, which outputs the candidate definition with the highest confidence score predicted by SDI-Extractor (Section \ref{sec:sdi}).
\item \textbf{CDM-S$k$,C$k'$}: The combined definition modeling model introduced in Section \ref{sec:generation}. S$k$ or C$k'$ is omitted when $k$ or $k'$ is equal to 0.
\end{itemize}

\begin{table}[tp!]
\begin{center}
\small
\setlength\tabcolsep{5pt}
\begin{tabular}{l|c|c|c}
\toprule
& \textbf{Precision} & \textbf{Recall} & \textbf{F1} \\
\hline
CNN & 91.84 & 90.66 & 91.25  \\
C-BLSTM & 91.59 & 88.93 & 90.24  \\
\textbf{SDI-Extractor} & \textbf{96.72} & \textbf{97.67} & \textbf{97.19}  \\
\bottomrule
\end{tabular}
\end{center}
\vspace{-1mm}
\caption{Results of definition extraction.}
\vspace{-3mm}
\label{table:clf}
\end{table}

\begin{table*}[tp!]
\begin{center}
\small
\setlength\tabcolsep{5pt} %
\begin{tabular}{l|cccc|cccc|cccc}
\toprule
 & \multicolumn{4}{c|}{\textbf{UJ-CS}} & \multicolumn{4}{c|}{\textbf{UJ-Math}} & \multicolumn{4}{c}{\textbf{UJ-Phy}} \\
 & \textbf{BL} & \textbf{R-L} & \textbf{MT} & \textbf{BS}  & \textbf{BL} & \textbf{R-L} & \textbf{MT} & \textbf{BS} & \textbf{BL} & \textbf{R-L} & \textbf{MT} & \textbf{BS} \\
\midrule 
\textbf{Extractive} & 15.62 & 29.41 & 16.41 & 79.02 & 13.04 & 25.95 & 13.88 & 75.97 & 9.75 & 20.30 & 12.48 & 75.27 \\
\hline
Gen (w/o context) & 8.31 & 28.02 & 12.83 & 77.97 & 6.89 & 28.50 & 10.97 & 76.45 & 5.28 & 25.75 & 10.57 & 76.88 \\
Gen (w/ context) & 8.76 & 30.00 & 13.15 & 78.73 & 10.00 & 31.18 & 12.67 & 77.24 & 8.71 & 29.68 & 12.94 & 78.37 \\
DMAS & 4.98 & 26.05 & 10.60 & 78.09 & 1.58 & 24.10 & 7.97 & 75.75 & 2.70 & 24.59 & 9.59 & 77.43 \\
\hline
\textbf{CDM-C$5$} & 12.26 & 29.90 & 14.55 & 79.09 & 12.54 & 32.17 & 14.10 & 78.22 & 9.36 & 28.87 & 13.26 & 78.62 \\
\hline
\textbf{CDM-S$1$} & 17.12 & 34.67 & 17.46 & 80.75 & 16.33 & 36.07 & 16.22 & 78.94 & 12.42 & 31.48 & 14.70 & 78.96 \\
\textbf{CDM-S$3$} & 19.08 & 35.48 & 18.44 & 81.16 & 19.35 & 38.56 & 17.88 & 79.90 & 16.54 & 34.54 & 17.00 & 80.03 \\
\textbf{CDM-S$5$} & \underline{20.21} & 35.98 & \underline{19.06} & 81.33 & 20.76 & \uline{39.87} & 18.63 & \uline{80.31} & 18.58 & 35.15 & 18.00 & 80.38 \\
\textbf{CDM-S$10$} & 19.27 & \underline{36.34} & 18.79 & \underline{81.51} & \uline{21.71} & \textbf{40.43} & \uline{19.28} & \textbf{80.68} & \uline{20.66} & \uline{36.92} & \uline{19.18} & \uline{81.03} \\
\hline
\textbf{CDM-S$5$,C$5$} & \textbf{22.66} & \textbf{38.12} & \textbf{20.30} & \textbf{82.00} & \textbf{23.22} & 39.39 & \textbf{19.61} & 80.30 & \textbf{20.84} & \textbf{37.66} & \textbf{19.26} & \textbf{81.18} \\
\bottomrule
\end{tabular}
\end{center}
\caption{Results of definition generation on automatic metrics. The best results are \textbf{bold} and second best ones are \underline{underlined}.}
\vspace{-2mm}
\label{table:dm}
\end{table*}

\begin{table}[tp!]
\begin{center}
\small
\setlength\tabcolsep{5pt}
\begin{tabular}{l|cccc}
\toprule
 & \textbf{BL} & \textbf{R-L} & \textbf{MT} & \textbf{BS} \\
\midrule 
\textbf{Extractive} & 8.75 & 17.79 & 12.32 & 74.68 \\
\hline
Gen (w/o context) & 13.13 & 31.75 & 13.30 & 79.31 \\
Gen (w/ context) & 12.50 & 31.50 & 13.86 & 79.54 \\
DMAS & 9.12 & 28.43 & 11.04 & 79.21  \\
BART NO SD & 10.68 & 30.89 & 13.18 & 79.19 \\
BART SD & 11.11 & 32.34 & 13.97 & 80.12 \\
\hline
\textbf{CDM-C$5$} & 13.50 & 32.19 & 15.00 & 80.24 \\
\hline
\textbf{CDM-S$1$} & 11.91 & 33.14 & 15.31 & 80.24 \\
\textbf{CDM-S$3$} & 17.97 & 35.60 & 17.23 & 81.30 \\
\textbf{CDM-S$5$} & 20.18 & 37.25 & 18.52 & 81.75 \\
\textbf{CDM-S$10$} & \underline{20.35} & \textbf{37.98} & \underline{19.22} & \textbf{82.19} \\
\hline
\textbf{CDM-S$5$,C$5$} & \textbf{20.55} & \underline{37.70} & \textbf{19.24} & \underline{81.98}  \\
\bottomrule
\end{tabular}
\end{center}
\caption{\revisejie{Results of definition generation on \textbf{Sci\&Med} \cite{august-2022-definition-complexity}.}}
\vspace{-2mm}
\label{table:results_new}
\end{table}

{\flushleft \textbf{Metrics}.}
\label{app:metrics}
For extraction, we use the standard precision, recall, and F1 scores to evaluate the performance.
For generation, we follow \citet{bevilacqua2020generationary} and apply several automatic metrics, including BLEU (BL)\footnote{The version implemented on \url{https://github.com/mjpost/sacrebleu.}} \cite{papineni2002bleu}, ROUGE-L (R-L) \cite{lin2004rouge}, METEOR (MT) \cite{banerjee2005meteor}, and BERTScore (BS) \cite{zhang2019bertscore}.
BLEU, ROUGE-L, and METEOR focus on measuring surface similarities between the generated definitions and the ground-truth definitions, and BERTScore is based on the similarities of contextual token embeddings.
The signature of BERTScore is: roberta-large-mnli L19 no-idf version=0.3.0(hug trans=2.8.0).
We also ask three human annotators \revisejie{(graduate students doing research on computational linguistics)} to evaluate the output definitions with a 1-5 rating scale used in \citet{ishiwatari2019learning}:
    1) completely wrong or self-definition; 
    2) correct topic with wrong information; 
    3) correct but incomplete; 
    4) small details missing; 
    5) correct.

{\flushleft \textbf{Implementation Details}.}
\label{app:implementation}
For SDI extraction, we adopt BERT-base-uncased from huggingface transformers framework \cite{wolf2020transformers}.
We apply the BertForSequenceClassification in huggingface (with a linear layer on top of the pooled output). We use the default hyperparameters and fine-tune the model using Adam \cite{kingma2014adam} with learning rate of $2 \times 10^{-6}$. All the layers of the BERT model are fine-tuned.
\revisejie{For the two baselines, we train the models on our data with the official implementation}.
For the extracted SDI, we exclude sentences from Wikipedia to avoid the models to see the ground truth.

For CDI extraction, following \citet{huang2021definition}, we use the built-in Elasticsearch-based Wikipedia search engine\footnote{\url{https://en.wikipedia.org/w/index.php?search}} to collect related core terms for jargon; 
and then, we extract the first sentence on the corresponding Wikipedia page as the definition of each related term.

For generation, we employ the fairseq library\footnote{\url{https://github.com/pytorch/fairseq/tree/master/examples/bart}} to build the BART-base generator and adopt the hyperparameters and settings as suggested in \citet{bevilacqua2020generationary}.
We set the learning rate as $5 \times 10^{-5}$
and use batch size of $1,024$ tokens, updating every $16$ iterations, with the number of warmup steps as $1,000$.
\revisejie{For all the datasets, we use the same trained SDI-extractor as described above to extract SDI.}
We adopt the default/suggested hyperparameters for the baselines.
\revisejie{We train and evaluate all the baselines and variants on the same train/valid/test split on NVIDIA Quadro RTX 5000 GPUs.}
The training of CDM can be finished in one hour.

\subsection{Definition Extraction}

Table \ref{table:clf} reports the results of definition extraction. We observe that SDI-Extractor outperforms baselines significantly and the performance is quite satisfactory (with an F1 score higher than $0.97$), 
which indicates our definition extractor can extract useful self-definitional information for jargon.

\subsection{Definition Generation}

We provide both quantitative and qualitative evaluations for definition generation.

\subsubsection{Automatic Evaluation}
Tables \ref{table:dm} and \ref{table:results_new} show the results on automatic metrics\footnote{For Table~\ref{table:results_new}, \citet{august-2022-definition-complexity} use BERT-base for BERTScore, while we use RoBERTa-large for BERTScore to be consistent with Table \ref{table:dm}.}.
We observe the proposed CDM model outperforms the SOTA baselines significantly.
Comparing Gen (w/ context) with Gen (w/o context), we find contexts (random sentences containing the target jargon) only have limited help with jargon definition modeling.
Besides, CDM-S$5$ outperforms CDM-S$3$, while CDM-S$3$ outperforms CDM-S$1$, which means the sentences extracted by SDI-Extractor can provide important definitional information.
Comparing CDM-C$5$ with Gen (w/ context) and Gen (w/o context), we can verify CDI is also helpful for definition generation, while the improvement is not as significant as the models with SDI, e.g., CDM-S$5$. 
Among all the models, CDM-S$5$,C$5$ usually achieves the best performance, which demonstrates the combination of SDI and CDI is the most significant for jargon definition modeling. 

An interesting finding is that our simple extractive model is comparable to the SOTA abstractive baselines \revisejie{(except for Table \ref{table:results_new}, because most of the definitions in the dataset are not complete sentences, e.g, ``\textit{the science of automatic control systems}'' for \textit{cybernetics}, while SDI-Extractor usually extracts complete sentences)}. 
We suppose this is because, compared to general words/phrases, jargons are more difficult to define without external knowledge. For instance, it is almost impossible for a model to generate the definition for \textit{twin prime} only with context ``\textit{proof of this conjecture would also imply the existence of an infinite number of \uline{twin primes}}'', while the definition can possibly be retrieved from the Web. The results also demonstrate that existing context-aware definition modeling systems are hard to handle jargon, while our proposed extraction-generation framework is quite practical for jargon definition modeling.

\begin{table}[tp!]
\begin{center}
\small
\begin{tabular}{l|c}
\toprule
 & \textbf{Score (1-5)} \\
\midrule 
\textbf{Extractive} & 3.57  \\
Gen (w/ context) & 2.34 \\
\textbf{CDM-S$1$} & 3.65 \\
\textbf{CDM-S$5$} & 3.99 \\
\textbf{CDM-S$5$,C$5$} & \textbf{4.04} \\
\bottomrule
\end{tabular}
\end{center}
\vspace{-1mm}
\caption{Averaged human annotated scores.}
\vspace{-2mm}
\label{table:human_evaluation}
\end{table}

\subsubsection{Human Evaluation}
\label{sec:human_eval}
We conduct human evaluation for the computer science field (UJ-CS). Specifically, we randomly sample 50 jargons from the test set, and ask three human annotators to evaluate the definitions produced by different models with the rating scale described in Section \ref{app:metrics}.
Table \ref{table:human_evaluation} reports the human evaluation results, where the average pairwise Cohen's $\kappa$ is 0.69 (good agreement). 
We observe the state-of-the-art baseline Gen (w/context) is difficult to generate reasonable definitions for jargon. 
In contrast, the proposed CDM-S$5$,C$5$ model can produce high-quality definitions in most cases (with a human-annotated score higher than $4$). The human evaluation results are also consistent with the automatic evaluation results presented in Table \ref{table:dm}.

\begin{figure}[tp!]
\centerline{\includegraphics[width=\linewidth]{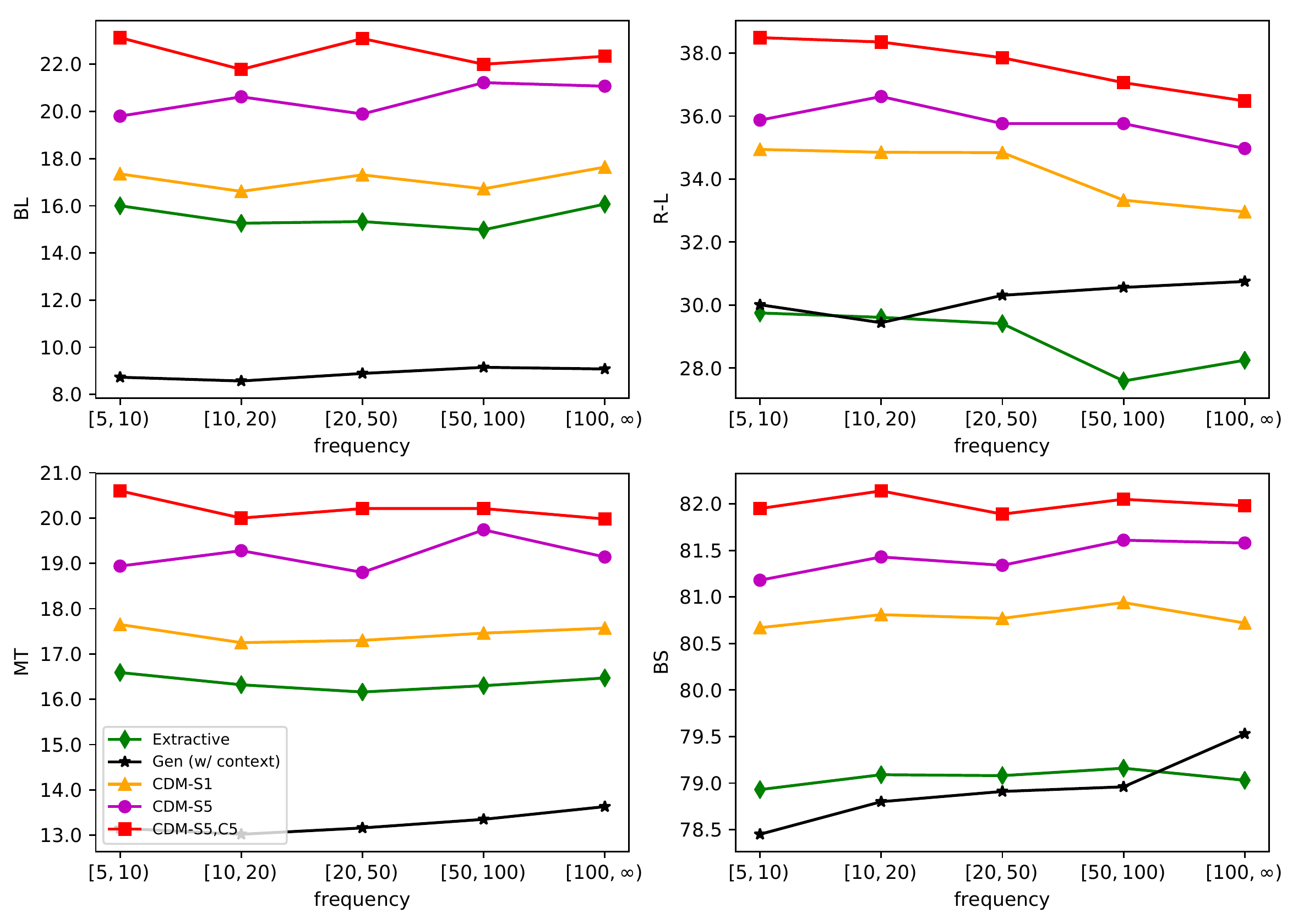}}
\vspace{-1mm}
\caption{Results of definition generation with respect to jargon frequency in Springer (author-assigned keywords). Best viewed in color.}
\vspace{-2mm}
\label{fig:results-freq}
\end{figure}

\begin{table*}[tp!]
\scriptsize
\setlength\tabcolsep{1.5pt}
\begin{tabular}{p{0.1\linewidth}|p{0.20\linewidth}|p{0.16\linewidth}|p{0.25\linewidth}|p{0.25\linewidth}}
\toprule
 & \textbf{Extractive} & \textbf{Gen (w/ context)} & \textbf{CDM-S$5$,C$5$} & \textbf{\textit{Gold}} \\
\midrule[1pt]
intelligent user interfaces & ACM IUI 2021 is the 26th annual premier international forum for reporting outstanding research and development on intelligent user interfaces. & An intelligent user interface (IUI) is a user interface (UI) that is designed to respond to the needs and wants of the user. & In computing, an intelligent user interface (UI) is a user interface that has been developed by a group of researchers interested in enabling natural human-computer interaction by combining techniques from computer vision, machine learning, computer graphics, human-computer interaction and psychology. & An intelligent user interface (Intelligent UI, IUI, or sometimes Interface Agent) is a user interface (UI) that involves some aspect of artificial intelligence (AI or computational intelligence). \\
\hline
wear leveling & The term preemptive wear leveling (PWL) has been used by Western Digital to describe their preservation technique used on hard disk drives (HDDs) designed for storing audio and video data. & Wear leveling is the process of reducing the wear of a Flash die. & Wear leveling is a technique used to increase the lifetime of a solid-state drive (SSD). & Wear leveling (also written as wear levelling) is a technique for prolonging the service life of some kinds of erasable computer storage media, such as flash memory, which is used in solid-state drives (SSDs) and USB flash drives, and phase-change memory. \\
\hline
gittins index & In applied mathematics, the "Gittins index" is a real scalar value associated to the state of a stochastic process with a reward function and with a probability of termination. & The Gittins index is a decision-making tool used in decision-making and project management. & In applied mathematics, the Gittins index is a real scalar value associated to the state of a stochastic process with a reward function and with a probability of termination. & The Gittins index is a measure of the reward that can be achieved through a given stochastic process with certain properties, namely: the process has an ultimate termination state and evolves with an option, at each intermediate state, of terminating. \\
\hline
reduplication & The term ?€?compensatory reduplication?€? refers to duplication that serves a phonological purpose. & In mathematics, reduplication is a generalization of the concept of reduplication. & Reduplication is the repetition of an entire word, word stem (root with one or more affixes), or root. & In linguistics, reduplication is a morphological process in which the root or stem of a word (or part of it) or even the whole word is repeated exactly or with a slight change. \\
\hline
power delay profile & The power delay profile of a channel represents the average power of the received signal in terms of the delay with respect to the first arrival path in multi-path transmission. & A power delay profile (PDP) is a measure of the time delay between the transmission and reception of a signal. & In telecommunications, the power delay profile (PDP) of a multipath channel represents the average power of the received signal in terms of the delay with respect to the first arrival path in multi-path transmission. & The power delay profile (PDP) gives the intensity of a signal received through a multipath channel as a function of time delay. \\
\bottomrule
\end{tabular}
\vspace{-1mm}
\caption{Sample of definitions produced by \textit{Extractive}, \textit{Gen (w/ context)}, and \textit{CDM-S$5$,C$5$}.}
\vspace{-2mm}
\label{table:example}
\end{table*}

\subsection{Sensitivity to Frequency}
\label{app:sensitivity}

To investigate the sensitivity of the models with respect to the popularity of jargon, we report the results according to jargon frequency in Figure \ref{fig:results-freq}. We observe that Generationary \cite{bevilacqua2020generationary} achieves slightly worse performance for less popular jargon on all metrics, while CDM performs well for low-frequency jargon, which indicates our framework can produce high-quality definitions for long-tail jargon. 
We suppose this is because, although long-tail jargon is less frequent, we can still extract useful definitional information from the entire Web and incorporate it for definition generation.

\subsection{Generation Examples and Error Analysis}
\label{app:example}

In Table \ref{table:example}, we show some sample outputs in the test set of three models: \textit{Extractive}, \textit{Gen (w/ context)}, and \textit{CDM-S$5$,C$5$}, with ground-truth definitions in Wikipedia (\textit{Gold}) as references. 

From the results, we observe although the extractive baseline can produce reasonable sentences, the output sentences may not be high-quality definitional sentences of the target jargon. For instance, the extracted sentence for \textit{wear leveling} in fact is the definition of \textit{preemptive wear leveling}. 
We also find Gen (w/ context) suffers severely from \emph{hallucinations}, i.e., generating irrelevant or contradicted facts. For instance, \textit{gittins index} is described as a decision-making tool instead of a measure/value, which is completely wrong.
This is mainly because the contexts of jargon may not provide sufficient knowledge to define jargon.
In contrast, the quality of definitions generated by CDM-S$5$,C$5$ is high--
all the generated definitions capture the main characteristics of the target jargon correctly.

{\flushleft \textbf{Error Analysis}.} To further understand the results and identify the remaining challenges, we analyze the human evaluation results. We find that errors could be introduced in either the extraction or the generation process. E.g., 1) for \textit{intelligent user interfaces} in Table \ref{table:example}, the top 1 sentence extracted by SDI-Extractor (``\textit{ACM IUI ... interfaces.}'') cannot provide meaningful knowledge to the generator. Although by incorporating other sentences,  CDM-S$5$,C$5$ can generate a reasonable definition, the definition still contains minor errors. 
2) For \textit{markup languages}, although SDI-Extractor extracts reasonable definitions (e.g., ``\textit{Markup languages are languages used by a computer to annotate a document.}''), the generator mistakenly synthesizes the SDI and CDI into ``\textit{A markup language is a series of tags mixed with plain text.}''
Nonetheless, compared to existing models that do not combine extraction and generation, CDM greatly reduces hallucination.

\vspace{-1mm}
\section{Discussion}
\vspace{-1mm}
\label{sec:discussion}

In this work, we focus on jargon definition modeling. The proposed framework can be further extended to general words/phrases in a context-aware setting \cite{gadetsky2018conditional}. For instance, to retrieve the definitional information, we can incorporate the context the target word/phrase used in. E.g., the BERT extractor can be trained with a modified encoding scheme: ``\textit{[CLS] $word/phrase$ [SEP] $context$ [DEF] $sentence$}''. Similarly, the generator can produce the final definition conditioned on the context. E.g., the input of the generator can be encoded as ``\textit{$word/phrase$ [SEP] $context$ [DEF] $sent_1$ [SEP] $sent_2$ ... [SEP] $sent_{k}$ [DEF] $sent'_1$ [SEP] $sent'_2$ ... [SEP] $sent'_{k'}$}''. 
Since our framework is modular, the BERT extractor and BART generator can also be replaced with more advanced language models.
It is also interesting to train the extractor and generator jointly or iteratively \cite{guu2020realm,lewis2020retrieval}. 
We keep the proposed model simple and leave context-aware combined definition modeling and more complicated combinations as future work. 

\vspace{-1mm}
\section{Conclusion}
\vspace{-1mm}

We present the first combination of definition extraction and definition generation.
We show that, by incorporating extracted self- and correlative definitional information, the generator can produce high-quality definitions for jargon.
Experimental results demonstrate the effectiveness of our framework, where the proposed method outperforms recent baselines by a large margin.
We also publish several datasets for jargon definition modeling.
In future work, we plan to improve our framework as discussed in Section \ref{sec:discussion} and apply our methods to construct several online domain dictionaries.

\section*{Limitations}

One limitation of this paper is that it does not consider the diversity of definitions. 
Definitions from different perspectives can facilitate a more comprehensive understanding.
For instance,
to define \textit{artificial intelligence}, we may relate it to or contrast it with other concepts, e.g., ``\textit{artificial intelligence refers to systems or machines that mimic human intelligence to perform tasks and can iteratively improve themselves based on the information they collect.}''
or
``\textit{artificial intelligence is intelligence demonstrated by machines, as opposed to the natural intelligence displayed by animals including humans.}''
Recent work starts to model the specificity and complexity for definition modeling \cite{huang2021definition,gardner2022definition}; however, the diversity of generative definitions is still limited. We believe our framework can benefit diversity since the generator has the potential to generate definitions with different styles by incorporating diverse definitional information extracted from the Web.

\section*{Acknowledgements}

We thank the reviewers for their constructive feedback.
This material is based upon work supported by the National Science Foundation IIS 16-19302 and IIS 16-33755, Zhejiang University ZJU Research 083650, IBM-Illinois Center for Cognitive Computing Systems Research (C3SR)-- a research collaboration as part of the IBM Cognitive Horizon Network, grants from eBay and Microsoft Azure, UIUC OVCR CCIL Planning Grant 434S34, UIUC CSBS Small Grant 434C8U, and UIUC New Frontiers Initiative. Any opinions, findings, and conclusions or recommendations expressed in this publication are those of the author(s) and do not necessarily reflect the views of the funding agencies.

\bibliography{anthology,custom}

\begin{thebibliography}{43}
\expandafter\ifx\csname natexlab\endcsname\relax\def\natexlab#1{#1}\fi

\bibitem[{Anke and Schockaert(2018)}]{anke2018syntactically}
Luis~Espinosa Anke and Steven Schockaert. 2018.
\newblock Syntactically aware neural architectures for definition extraction.
\newblock In \emph{Proceedings of the 2018 Conference of the North American
  Chapter of the Association for Computational Linguistics: Human Language
  Technologies, Volume 2 (Short Papers)}, pages 378--385.

\bibitem[{August et~al.(2022)August, Reinecke, and
  Smith}]{august-2022-definition-complexity}
Tal August, Katharina Reinecke, and Noah~A Smith. 2022.
\newblock Generating scientific definitions with controllable complexity.
\newblock In \emph{ACL}.

\bibitem[{Banerjee and Lavie(2005)}]{banerjee2005meteor}
Satanjeev Banerjee and Alon Lavie. 2005.
\newblock Meteor: An automatic metric for mt evaluation with improved
  correlation with human judgments.
\newblock In \emph{Proceedings of the acl workshop on intrinsic and extrinsic
  evaluation measures for machine translation and/or summarization}, pages
  65--72.

\bibitem[{Ben~Abacha and Demner-Fushman(2019)}]{ben2019question}
Asma Ben~Abacha and Dina Demner-Fushman. 2019.
\newblock A question-entailment approach to question answering.
\newblock \emph{BMC bioinformatics}, 20(1):1--23.

\bibitem[{Bevilacqua et~al.(2020)Bevilacqua, Maru, and
  Navigli}]{bevilacqua2020generationary}
Michele Bevilacqua, Marco Maru, and Roberto Navigli. 2020.
\newblock Generationary or:“how we went beyond word sense inventories and
  learned to gloss”.
\newblock In \emph{Proceedings of the 2020 Conference on Empirical Methods in
  Natural Language Processing (EMNLP)}, pages 7207--7221.

\bibitem[{Brown et~al.(2020)Brown, Mann, Ryder, Subbiah, Kaplan, Dhariwal,
  Neelakantan, Shyam, Sastry, Askell, Agarwal, Herbert-Voss, Krueger, Henighan,
  Child, Ramesh, Ziegler, Wu, Winter, Hesse, Chen, Sigler, Litwin, Gray, Chess,
  Clark, Berner, McCandlish, Radford, Sutskever, and
  Amodei}]{NEURIPS2020_1457c0d6}
Tom Brown, Benjamin Mann, Nick Ryder, Melanie Subbiah, Jared~D Kaplan, Prafulla
  Dhariwal, Arvind Neelakantan, Pranav Shyam, Girish Sastry, Amanda Askell,
  Sandhini Agarwal, Ariel Herbert-Voss, Gretchen Krueger, Tom Henighan, Rewon
  Child, Aditya Ramesh, Daniel Ziegler, Jeffrey Wu, Clemens Winter, Chris
  Hesse, Mark Chen, Eric Sigler, Mateusz Litwin, Scott Gray, Benjamin Chess,
  Jack Clark, Christopher Berner, Sam McCandlish, Alec Radford, Ilya Sutskever,
  and Dario Amodei. 2020.
\newblock Language models are few-shot learners.
\newblock In \emph{Advances in Neural Information Processing Systems},
  volume~33, pages 1877--1901. Curran Associates, Inc.

\bibitem[{Bullock et~al.(2019)Bullock, Col{\'o}n~Amill, Shulman, and
  Dixon}]{bullock2019jargon}
Olivia~M Bullock, Daniel Col{\'o}n~Amill, Hillary~C Shulman, and Graham~N
  Dixon. 2019.
\newblock Jargon as a barrier to effective science communication: Evidence from
  metacognition.
\newblock \emph{Public Understanding of Science}, 28(7):845--853.

\bibitem[{Butterfield et~al.(2016)Butterfield, Ngondi, and
  Kerr}]{butterfield2016dictionary}
Andrew Butterfield, Gerard~Ekembe Ngondi, and Anne Kerr. 2016.
\newblock \emph{A dictionary of computer science}.
\newblock Oxford University Press.

\bibitem[{Cui et~al.(2004)Cui, Kan, and Chua}]{cui2004unsupervised}
Hang Cui, Min-Yen Kan, and Tat-Seng Chua. 2004.
\newblock Unsupervised learning of soft patterns for generating definitions
  from online news.
\newblock In \emph{Proceedings of the 13th international conference on World
  Wide Web}, pages 90--99.

\bibitem[{Devlin et~al.(2019)Devlin, Chang, Lee, and
  Toutanova}]{devlin2019bert}
Jacob Devlin, Ming-Wei Chang, Kenton Lee, and Kristina Toutanova. 2019.
\newblock Bert: Pre-training of deep bidirectional transformers for language
  understanding.
\newblock In \emph{Proceedings of the 2019 Conference of the North American
  Chapter of the Association for Computational Linguistics: Human Language
  Technologies, Volume 1 (Long and Short Papers)}, pages 4171--4186.

\bibitem[{Fahmi and Bouma(2006)}]{fahmi2006learning}
Ismail Fahmi and Gosse Bouma. 2006.
\newblock Learning to identify definitions using syntactic features.
\newblock In \emph{Proceedings of the Workshop on Learning Structured
  Information in Natural Language Applications}.

\bibitem[{Gadetsky et~al.(2018)Gadetsky, Yakubovskiy, and
  Vetrov}]{gadetsky2018conditional}
Artyom Gadetsky, Ilya Yakubovskiy, and Dmitry Vetrov. 2018.
\newblock Conditional generators of words definitions.
\newblock In \emph{Proceedings of the 56th Annual Meeting of the Association
  for Computational Linguistics (Volume 2: Short Papers)}, pages 266--271.

\bibitem[{Gardner et~al.(2022)Gardner, Khan, and Hung}]{gardner2022definition}
Noah Gardner, Hafiz Khan, and Chih-Cheng Hung. 2022.
\newblock Definition modeling: literature review and dataset analysis.
\newblock \emph{Applied Computing and Intelligence}, 2(1):83--98.

\bibitem[{Gormley and Tong(2015)}]{gormley2015elasticsearch}
Clinton Gormley and Zachary Tong. 2015.
\newblock \emph{Elasticsearch: the definitive guide: a distributed real-time
  search and analytics engine}.
\newblock " O'Reilly Media, Inc.".

\bibitem[{Guu et~al.(2020)Guu, Lee, Tung, Pasupat, and Chang}]{guu2020realm}
Kelvin Guu, Kenton Lee, Zora Tung, Panupong Pasupat, and Ming-Wei Chang. 2020.
\newblock Realm: Retrieval-augmented language model pre-training.
\newblock \emph{arXiv preprint arXiv:2002.08909}.

\bibitem[{Huang et~al.(2021{\natexlab{a}})Huang, Kajiwara, and
  Arase}]{huang2021definition}
Han Huang, Tomoyuki Kajiwara, and Yuki Arase. 2021{\natexlab{a}}.
\newblock Definition modelling for appropriate specificity.
\newblock In \emph{Proceedings of the 2021 Conference on Empirical Methods in
  Natural Language Processing}, pages 2499--2509.

\bibitem[{Huang et~al.(2021{\natexlab{b}})Huang, Chang, Xiong, and
  Hwu}]{huang2021measuring}
Jie Huang, Kevin Chen-Chuan Chang, Jinjun Xiong, and Wen-mei Hwu.
  2021{\natexlab{b}}.
\newblock Measuring fine-grained domain relevance of terms: A hierarchical
  core-fringe approach.
\newblock In \emph{Proceedings of the 59th Annual Meeting of the Association
  for Computational Linguistics and the 11th International Joint Conference on
  Natural Language Processing}.

\bibitem[{Huang et~al.(2022)Huang, Chang, Xiong, and Hwu}]{huang2022open}
Jie Huang, Kevin Chen-Chuan Chang, Jinjun Xiong, and Wen-mei Hwu. 2022.
\newblock Open relation modeling: Learning to define relations between
  entities.
\newblock In \emph{Findings of the Association for Computational Linguistics:
  ACL 2022}.

\bibitem[{Ishiwatari et~al.(2019)Ishiwatari, Hayashi, Yoshinaga, Neubig, Sato,
  Toyoda, and Kitsuregawa}]{ishiwatari2019learning}
Shonosuke Ishiwatari, Hiroaki Hayashi, Naoki Yoshinaga, Graham Neubig, Shoetsu
  Sato, Masashi Toyoda, and Masaru Kitsuregawa. 2019.
\newblock Learning to describe unknown phrases with local and global contexts.
\newblock In \emph{Proceedings of the 2019 Conference of the North American
  Chapter of the Association for Computational Linguistics: Human Language
  Technologies, Volume 1 (Long and Short Papers)}, pages 3467--3476.

\bibitem[{Jin et~al.(2013)Jin, Kan, Ng, and He}]{jin2013mining}
Yiping Jin, Min-Yen Kan, Jun~Ping Ng, and Xiangnan He. 2013.
\newblock Mining scientific terms and their definitions: A study of the acl
  anthology.
\newblock In \emph{Proceedings of the 2013 Conference on Empirical Methods in
  Natural Language Processing}, pages 780--790.

\bibitem[{Kang et~al.(2020)Kang, Head, Sidhu, Lo, Weld, and
  Hearst}]{kang2020document}
Dongyeop Kang, Andrew Head, Risham Sidhu, Kyle Lo, Daniel~S Weld, and Marti~A
  Hearst. 2020.
\newblock Document-level definition detection in scholarly documents: Existing
  models, error analyses, and future directions.
\newblock In \emph{Proceedings of the First Workshop on Scholarly Document
  Processing}, pages 196--206.

\bibitem[{Kingma and Ba(2015)}]{kingma2014adam}
Diederik~P Kingma and Jimmy Ba. 2015.
\newblock Adam: A method for stochastic optimization.
\newblock \emph{Proceedings of the 3rd International Conference on Learning
  Representations}.

\bibitem[{Klavans and Muresan(2001)}]{klavans2001evaluation}
Judith~L Klavans and Smaranda Muresan. 2001.
\newblock Evaluation of the definder system for fully automatic glossary
  construction.
\newblock In \emph{Proceedings of the AMIA Symposium}, page 324. American
  Medical Informatics Association.

\bibitem[{Lewis et~al.(2020{\natexlab{a}})Lewis, Liu, Goyal, Ghazvininejad,
  Mohamed, Levy, Stoyanov, and Zettlemoyer}]{lewis2020bart}
Mike Lewis, Yinhan Liu, Naman Goyal, Marjan Ghazvininejad, Abdelrahman Mohamed,
  Omer Levy, Veselin Stoyanov, and Luke Zettlemoyer. 2020{\natexlab{a}}.
\newblock Bart: Denoising sequence-to-sequence pre-training for natural
  language generation, translation, and comprehension.
\newblock In \emph{Proceedings of the 58th Annual Meeting of the Association
  for Computational Linguistics}, pages 7871--7880.

\bibitem[{Lewis et~al.(2020{\natexlab{b}})Lewis, Perez, Piktus, Petroni,
  Karpukhin, Goyal, K{\"u}ttler, Lewis, Yih, Rockt{\"a}schel
  et~al.}]{lewis2020retrieval}
Patrick Lewis, Ethan Perez, Aleksandra Piktus, Fabio Petroni, Vladimir
  Karpukhin, Naman Goyal, Heinrich K{\"u}ttler, Mike Lewis, Wen-tau Yih, Tim
  Rockt{\"a}schel, et~al. 2020{\natexlab{b}}.
\newblock Retrieval-augmented generation for knowledge-intensive nlp tasks.
\newblock \emph{arXiv preprint arXiv:2005.11401}.

\bibitem[{Li et~al.(2020)Li, Bao, Huang, Dai, and Jiajun}]{li2020explicit}
Jiahuan Li, Yu~Bao, Shujian Huang, Xinyu Dai, and CHEN Jiajun. 2020.
\newblock Explicit semantic decomposition for definition generation.
\newblock In \emph{Proceedings of the 58th Annual Meeting of the Association
  for Computational Linguistics}, pages 708--717.

\bibitem[{Lin(2004)}]{lin2004rouge}
Chin-Yew Lin. 2004.
\newblock Rouge: A package for automatic evaluation of summaries.
\newblock In \emph{Text summarization branches out}, pages 74--81.

\bibitem[{Liu et~al.(2021)Liu, Wang, Gu, Zhang, Zhang, and
  Wang}]{liu2021graphine}
Zequn Liu, Shukai Wang, Yiyang Gu, Ruiyi Zhang, Ming Zhang, and Sheng Wang.
  2021.
\newblock Graphine: A dataset for graph-aware terminology definition
  generation.
\newblock In \emph{Proceedings of the 2021 Conference on Empirical Methods in
  Natural Language Processing}, pages 3453--3463.

\bibitem[{Mickus et~al.(2019)Mickus, Paperno, and Constant}]{mickus2019mark}
Timothee Mickus, D~Paperno, and Mathieu Constant. 2019.
\newblock Mark my word: A sequence-to-sequence approach to definition modeling.
\newblock In \emph{Proceedings of The First NLPL Workshop on Deep Learning for
  Natural Language Processing}, page~1. Link{\"o}ping University Electronic
  Press.

\bibitem[{Noraset et~al.(2017)Noraset, Liang, Birnbaum, and
  Downey}]{noraset2017definition}
Thanapon Noraset, Chen Liang, Larry Birnbaum, and Doug Downey. 2017.
\newblock Definition modeling: Learning to define word embeddings in natural
  language.
\newblock In \emph{Proceedings of the AAAI Conference on Artificial
  Intelligence}, volume~31.

\bibitem[{Papineni et~al.(2002)Papineni, Roukos, Ward, and
  Zhu}]{papineni2002bleu}
Kishore Papineni, Salim Roukos, Todd Ward, and Wei-Jing Zhu. 2002.
\newblock Bleu: a method for automatic evaluation of machine translation.
\newblock In \emph{Proceedings of the 40th annual meeting of the Association
  for Computational Linguistics}, pages 311--318.

\bibitem[{Radford et~al.(2019)Radford, Wu, Child, Luan, Amodei, and
  Sutskever}]{radford2019language}
Alec Radford, Jeffrey Wu, Rewon Child, David Luan, Dario Amodei, and Ilya
  Sutskever. 2019.
\newblock Language models are unsupervised multitask learners.
\newblock \emph{OpenAI blog}, 1(8):9.

\bibitem[{Raffel et~al.(2020)Raffel, Shazeer, Roberts, Lee, Narang, Matena,
  Zhou, Li, and Liu}]{raffel2020exploring}
Colin Raffel, Noam Shazeer, Adam Roberts, Katherine Lee, Sharan Narang, Michael
  Matena, Yanqi Zhou, Wei Li, and Peter~J Liu. 2020.
\newblock Exploring the limits of transfer learning with a unified text-to-text
  transformer.
\newblock \emph{Journal of Machine Learning Research}, 21:1--67.

\bibitem[{Reid et~al.(2020)Reid, Marrese-Taylor, and Matsuo}]{reid2020vcdm}
Machel Reid, Edison Marrese-Taylor, and Yutaka Matsuo. 2020.
\newblock Vcdm: Leveraging variational bi-encoding and deep contextualized word
  representations for improved definition modeling.
\newblock In \emph{Proceedings of the 2020 Conference on Empirical Methods in
  Natural Language Processing (EMNLP)}, pages 6331--6344.

\bibitem[{Vanetik et~al.(2020)Vanetik, Litvak, Shevchuk, and
  Reznik}]{vanetik2020automated}
Natalia Vanetik, Marina Litvak, Sergey Shevchuk, and Lior Reznik. 2020.
\newblock Automated discovery of mathematical definitions in text.
\newblock In \emph{Proceedings of The 12th Language Resources and Evaluation
  Conference}, pages 2086--2094.

\bibitem[{Veyseh et~al.(2020)Veyseh, Dernoncourt, Dou, and
  Nguyen}]{veyseh2020joint}
Amir Veyseh, Franck Dernoncourt, Dejing Dou, and Thien Nguyen. 2020.
\newblock A joint model for definition extraction with syntactic connection and
  semantic consistency.
\newblock In \emph{Proceedings of the AAAI Conference on Artificial
  Intelligence}, volume~34, pages 9098--9105.

\bibitem[{Washio et~al.(2019)Washio, Sekine, and Kato}]{washio2019bridging}
Koki Washio, Satoshi Sekine, and Tsuneaki Kato. 2019.
\newblock Bridging the defined and the defining: Exploiting implicit lexical
  semantic relations in definition modeling.
\newblock In \emph{Proceedings of the 2019 Conference on Empirical Methods in
  Natural Language Processing and the 9th International Joint Conference on
  Natural Language Processing (EMNLP-IJCNLP)}, pages 3521--3527.

\bibitem[{Westerhout(2009)}]{westerhout2009definition}
Eline Westerhout. 2009.
\newblock Definition extraction using linguistic and structural features.
\newblock In \emph{Proceedings of the 1st Workshop on Definition Extraction},
  pages 61--67.

\bibitem[{Wolf et~al.(2020)Wolf, Chaumond, Debut, Sanh, Delangue, Moi, Cistac,
  Funtowicz, Davison, Shleifer et~al.}]{wolf2020transformers}
Thomas Wolf, Julien Chaumond, Lysandre Debut, Victor Sanh, Clement Delangue,
  Anthony Moi, Pierric Cistac, Morgan Funtowicz, Joe Davison, Sam Shleifer,
  et~al. 2020.
\newblock Transformers: State-of-the-art natural language processing.
\newblock In \emph{Proceedings of the 2020 Conference on Empirical Methods in
  Natural Language Processing: System Demonstrations}, pages 38--45.

\bibitem[{Yang et~al.(2020)Yang, Kong, Chen, Liu, Fan, and
  Yang}]{yang2020incorporating}
Liner Yang, Cunliang Kong, Yun Chen, Yang Liu, Qinan Fan, and Erhong Yang.
  2020.
\newblock Incorporating sememes into chinese definition modeling.
\newblock \emph{IEEE/ACM Transactions on Audio, Speech, and Language
  Processing}, 28:1669--1677.

\bibitem[{Zhang et~al.(2022)Zhang, Roller, Goyal, Artetxe, Chen, Chen, Dewan,
  Diab, Li, Lin et~al.}]{zhang2022opt}
Susan Zhang, Stephen Roller, Naman Goyal, Mikel Artetxe, Moya Chen, Shuohui
  Chen, Christopher Dewan, Mona Diab, Xian Li, Xi~Victoria Lin, et~al. 2022.
\newblock Opt: Open pre-trained transformer language models.
\newblock \emph{arXiv preprint arXiv:2205.01068}.

\bibitem[{Zhang et~al.(2019)Zhang, Kishore, Wu, Weinberger, and
  Artzi}]{zhang2019bertscore}
Tianyi Zhang, Varsha Kishore, Felix Wu, Kilian~Q Weinberger, and Yoav Artzi.
  2019.
\newblock Bertscore: Evaluating text generation with bert.
\newblock In \emph{International Conference on Learning Representations}.

\bibitem[{Zheng et~al.(2021)Zheng, Dai, Li, Liu, Sui, Chang, and
  Liu}]{zheng2021decompose}
Hua Zheng, Damai Dai, Lei Li, Tianyu Liu, Zhifang Sui, Baobao Chang, and Yang
  Liu. 2021.
\newblock Decompose, fuse and generate: A formation-informed method for chinese
  definition generation.
\newblock In \emph{Proceedings of the 2021 Conference of the North American
  Chapter of the Association for Computational Linguistics: Human Language
  Technologies}, pages 5524--5531.

\end{thebibliography}
\bibliographystyle{acl_natbib}

\end{document}